%% file: bets.tex
\documentclass{article} 
\usepackage{iclr2027_conference,times}

\PassOptionsToPackage{hyphens}{url} 
\usepackage{hyperref}
\usepackage{url}
\usepackage{graphicx}

\usepackage{algorithm}
\usepackage{algorithmic}

\usepackage{booktabs}
\usepackage{threeparttable}
\usepackage{subcaption}
\usepackage{amssymb}
\usepackage{amsmath}
\usepackage{mathtools}
\usepackage{xcolor} 

\usepackage{tikz}
\usetikzlibrary{arrows.meta, positioning, calc, fit, backgrounds}

\usepackage{amsthm}
\theoremstyle{plain}

\theoremstyle{definition}

\theoremstyle{remark}

\input{common/definitions}

\input{fig/fig-styles}

\title{Belief Embedding Tree Search}

\author{%
\textbf{Christopher Solinas\textsuperscript{1},
Radovan Halu\v{s}ka\textsuperscript{3},
Martin Schmid\textsuperscript{3,4},} \\
\textbf{Michael Buro\textsuperscript{1},
Nathan R. Sturtevant\textsuperscript{1,2},
Michael Bowling\textsuperscript{1,2}} \\[3pt]
\textsuperscript{1}University of Alberta,
\textsuperscript{2}Alberta Machine Intelligence Institute (Amii) \\
\textsuperscript{3}Charles University, Prague,
\textsuperscript{4}EquiLibre Technologies, Inc. \\[3pt]
\texttt{\{solinas,mburo,nathanst,mbowling\}@ualberta.ca} \\
\texttt{\{haluskr,schmidm\}@kam.mff.cuni.cz}
}

\newcommand{\preprintcopy}[1][Preprint. Under review.]{%
  \iclrfinalcopy
  \let\iclrmaketitle\maketitle
  \renewcommand{\maketitle}{\iclrmaketitle\lhead{#1}}%
}
\preprintcopy

\begin{document}

\maketitle

\begin{abstract}
Beliefs in a Partially Observable Markov Decision Process (POMDP) are posterior distributions over the hidden state given the agent's history of actions and observations.
They summarize everything the history says about the state, so planning in a POMDP is planning in a Markov decision process (MDP) over beliefs.
Belief MDPs have continuous state spaces even for finite POMDPs, and exact planning in them is intractable in general.
Planners must therefore approximate beliefs and estimate values.
Monte Carlo planners do both inside a search tree, with an approximate belief at every node and a value estimate at every leaf.
Approximations usually consist of sets of sampled states, with leaf values coming from rollouts, hand-designed bounds, or learned networks.
Policies and values computed from these approximations are limited by approximation quality, and prior methods with learned networks rely on hand-chosen summaries of the sampled states to compute them.
A belief representation should be computable from the parent node's representation using the POMDP's dynamics and it should let the search sample from it, but it should also be learned so that it provides the information needed to condition policies and value functions.
We present Belief Embedding Tree Search (BETS), a Monte Carlo planner that meets these requirements by representing each node's belief with a generative distribution embedding (GDE) while updating it through the POMDP dynamics.
BETS trains its belief model alongside its policy and value networks through experience, with the realized hidden state of the POMDP as the belief model's training target.
We evaluate BETS on six configurations of RockSample, LightDark, and PocMan against BetaZero and three planners without learned components.
BETS is competitive or leading on all six, while every baseline falls far behind on at least one.
Where beliefs are highest-dimensional, it outperforms them by a large margin, even without search.
\end{abstract}

\input{sections/intro}
\input{sections/bg}
\input{sections/bets}
\input{sections/training}
\input{sections/experiments}
\input{sections/limitations}
\input{sections/conclusion}

\subsection*{AI Use Statement}
Large language models assisted the authors with writing and coding.
They were used to help the authors refine writing and help implement and test the experiment pipeline. The authors revised and verified all LLM work carefully.

\subsection*{Reproducibility Statement}
All source code for BETS is provided as supplementary material and will be made publicly available on GitHub upon publication.
Appendix~\ref{app:hyperparameters} gives the hyperparameter settings of BETS and of every baseline.

\bibliography{bets}
\bibliographystyle{iclr2027_conference}

\appendix

\input{appendix/domains}
\input{appendix/hyperparameters}
\input{appendix/ar-conditioning}
\input{appendix/wallclock}

\end{document}

%% file: common/definitions.tex
\renewcommand{\cite}[1]{{\color{red}USE CITEP OR CITET}}

\newcommand{\mc}{\mathcal}
\newcommand{\mb}{\mathbb}
\newcommand{\reals}{\mb R}
\newcommand{\R}{\reals}
\newcommand{\E}{\mathbb{E}}

\newcommand{\defeq}{\ensuremath{\stackrel{\rm def}{=}}}
\newcommand{\dist}{\Delta}

\newcommand{\state}{x}
\newcommand{\statespace}{X}

\newcommand{\transitionfunc}{T}

\newcommand{\obs}{y}
\newcommand{\obsspace}{Y}
\newcommand{\obsfunc}{\mc Y}

\newcommand{\jointpolicy}{\pi}

\newcommand{\action}{a}
\newcommand{\actionspace}{A}

\newcommand{\discount}{\gamma}
\newcommand{\rewardfunc}{R}

\newcommand{\env}{G}

\newcommand{\initialstatedist}{b_0}

\NewDocumentCommand{\kernel}{o o}{K_{\IfValueTF{#1}{#1}{\action},\IfValueTF{#2}{#2}{\obs}}}

\newcommand{\history}{h}
\newcommand{\statetime}[1]{\state^{(#1)}}
\newcommand{\actiontime}[1]{\action^{(#1)}}

\newcommand{\historyset}{H}

\newcommand{\obstime}[1]{\obs^{(#1)}}

\newcommand{\bdistparams}{\theta}

\newcommand{\bembed}{\bdistparams}

\newcommand{\embedfuncparams}{\phi}

\newcommand{\embedfunc}{\mc E_\embedfuncparams}

\newcommand{\generator}{\mc G_\flowparams}

\newcommand{\flowparams}{\psi}

\newcommand{\shortloss}{\mc L(\embedfuncparams, \flowparams)}

\newcommand{\particle}{\state_i}

\newcommand{\particleweight}{w_i}

\newcommand{\T}{t_{\text{max}}}

\NewDocumentCommand{\markovstate}{o}{
  \history\IfValueT{#1}{_{#1}}
}

\newcommand{\actionvalue}{Q}
\newcommand{\returnt}{\sum_{t^\prime = t}^{\T - 1}\discount^{t^\prime - t}\,\rewardfunc(\statetime{t^\prime},\actiontime{t^\prime},\statetime{t^\prime+1})} 
\newcommand{\nextparticle}{\state_i^\prime}

\newcommand{\weightedparticleset}{S}

\newcommand{\policynet}{\pi_\omega}
\newcommand{\obsfeat}{\xi} 
\newcommand{\policynetdef}{\policynet : \reals^{d_{\text{emb}}} \times \reals^{d_\obsfeat} \rightarrow \dist \actionspace}
\newcommand{\valuenet}{V_\eta}
\newcommand{\valuenetdef}{\valuenet : \reals^{d_{\text{emb}}} \times \reals^{d_\obsfeat} \rightarrow \reals}
\newcommand{\policyprior}{\rho}
\newcommand{\priormix}{m}
\newcommand{\unif}{u}
\newcommand{\pibar}{\bar{\pi}}
\newcommand{\vbar}{\bar{v}}
\newcommand{\readtemp}{\tau}
\newcommand{\kl}[2]{\mathrm{KL}\!\left(#1 \,\|\, #2\right)}

%% file: fig/fig-styles.tex
\definecolor{mplBluesA}{HTML}{D6E6F4}
\definecolor{mplBluesB}{HTML}{9ECAE1}
\definecolor{mplBluesC}{HTML}{6BAED6}
\definecolor{mplBluesD}{HTML}{3182BD}
\definecolor{mplOrgA}{HTML}{FEE6C5}
\definecolor{mplOrgB}{HTML}{FDD09E}
\definecolor{mplOrgC}{HTML}{FDB863}
\definecolor{mplOrgD}{HTML}{F59A35}
\definecolor{mplObs}{HTML}{FDD9A6}

\newcommand{\embedstrip}[5][3.4mm]{%
  \begin{tikzpicture}[baseline=-0.6ex]
    \pgfmathsetlengthmacro{\h}{1.5*#1}%
    \fill[#2] (0,0)     rectangle (#1,\h);
    \fill[#3] (#1,0)    rectangle (2*#1,\h);
    \fill[#4] (2*#1,0)  rectangle (3*#1,\h);
    \fill[#5] (3*#1,0)  rectangle (4*#1,\h);
    \foreach \i in {1,2,3}{\draw[white, line width=0.7pt] (\i*#1,0) -- (\i*#1,\h);}%
    \draw[gray!45, line width=0.4pt] (0,0) rectangle (4*#1,\h);
  \end{tikzpicture}%
}
\newcommand{\bstrip}[1][3.4mm]{\embedstrip[#1]{mplBluesA}{mplBluesB}{mplBluesC}{mplBluesD}}
\newcommand{\ostrip}[1][3.4mm]{\embedstrip[#1]{mplOrgA}{mplOrgB}{mplOrgC}{mplOrgD}}

\tikzset{
  >=Latex,
  embpanel/.style={draw, rounded corners=2pt, thick, fill=white, inner sep=2.5pt, align=center},
  scatterpanel/.style={draw, rounded corners=2pt, thick, fill=white,
                       minimum width=17mm, minimum height=15mm, inner sep=2pt},
  treenode/.style={draw, rounded corners=2pt, semithick, fill=white, inner sep=1.5pt},
  treenew/.style={treenode, draw=orange!80!black, line width=0.9pt},
  oplabel/.style={font=\scriptsize, align=center, inner sep=1pt},
  linelabel/.style={font=\scriptsize\itshape, text=gray!55!black, inner sep=1pt},
  obsnode/.style={draw=orange!70!black, rounded corners=1.5pt, thick, fill=mplObs,
                  minimum width=7mm, minimum height=5.5mm, font=\small},
  arr/.style={->, thick, shorten >=1pt, shorten <=1pt},
  treeedge/.style={semithick, gray!55!black},
  seledge/.style={->, thick, orange!80!black, dashed},
}

%% file: sections/intro.tex
\section{Introduction} \label{sec:intro}
Agents planning in a Partially Observable Markov Decision Process (POMDP) must act on a \emph{belief}, the posterior distribution over hidden states given their history of actions and observations.
The action an agent takes should depend on the belief rather than on any single state, so policies map beliefs to actions and value functions map beliefs to expected returns.
Computing them exactly is intractable in general, even when the state, action, and observation spaces are finite \citep{papadimitriou1987complexity}.
Monte Carlo planners instead search forward from the agent's current history by simulating the POMDP's dynamics and observation model \citep{silver2010monte,somani2013despot}.
These simulations draw states from a representation of the belief at every node of the search tree and stop at leaves that need value estimates.

Monte Carlo planners typically represent each node's belief with a set of sampled states and differ in how they maintain that set and how they estimate leaf values.
At each node POMCP and POMCPOW keep the states that simulations carried into it and estimate leaf values with rollouts of a default policy \citep{silver2010monte,sunberg2018online}.
DESPOT and AdaOPS search a fixed set of sampled scenarios and estimate leaf values with upper and lower bounds \citep{somani2013despot,wu2021adaptive}.
The accuracy of these leaf values depends on the rollout policies and bounds, which are either designed for each domain or loose.
Maintaining a set of sampled states also carries the weaknesses of particle filters.
The number of states needed for an accurate approximation grows rapidly with the size of the state space \citep{thrun2002particle,snyder2008obstacles}, and repeated resampling also collapses the set onto a few distinct states, a failure called \emph{impoverishment} \citep{sokota2022fine}.

Policy and value networks trained on experience generated by the planner can replace rollouts and bounds \citep{silver2018general}.
The networks of BetaZero \citep{moss2023betazero} condition on a hand-chosen summary of the particle set, the mean and standard deviation of its features.
This summary discards much of what distinguishes one belief from another, such as its support and the dependence between its variables.
BetaZero also keeps a particle set at every node of its tree, so it carries the same weaknesses of particle filters.
A learned belief representation that does not collapse under repeated updates would address both, provided updating it and sampling from it remain tractable.

\citet{solinas2025neural} proposed Neural Bayesian Filtering (NBF), which represents beliefs as generative distribution embeddings (GDEs) \citep{fishman2025generative} and approximates their Bayesian update after actions and observations with a forward pass through the GDE's generator and encoder.
At each step, NBF generates particles from the previous belief's embedding, propagates and reweights them under the POMDP's dynamics and likelihood, and embeds the weighted set as the next belief.
Every update therefore draws a fresh particle set, so impoverishment from resampling does not accumulate across steps.
NBF's conditional generator is trained to sample from the belief the embedding represents, so the search can generate the states it needs.
If the embedding is accurate it is also a sufficient statistic of the history \citep{fishman2025generative}, so policies and value functions can condition on it in place of the belief.

We introduce Belief Embedding Tree Search (BETS), a Monte Carlo planner that searches over a tree of embeddings.
Every node stores one embedding, expanding a leaf is one NBF update, and the policy and value networks that guide the search condition on the embedding, the same representation of the belief that the generator samples from.
Hidden states in many discrete POMDPs are configurations of coupled variables, so BETS adds a \emph{variable-coupling} stream to the GDE that pools a joint feature of every pair of variables over the particle set.
The belief model and both networks train together on episodes collected by running BETS, and the belief model's target is the hidden state the environment realizes at each step \citep{solinas2025neural}.

We evaluate BETS against BetaZero, POMCPOW, DESPOT, and AdaOPS on six configurations of RockSample, LightDark, and PocMan \citep{smith2004heuristic,platt2010belief,silver2010monte}.
Each baseline falls far below the best return in at least one setting, while BETS is competitive or leading on all six.
On RockSample(20,20) and PocMan, the configurations with the highest-dimensional beliefs, BETS outperforms every baseline by a large margin, even when it acts with its policy network alone.
On PocMan, value networks predict the return more accurately from embeddings than from summary statistics of a particle filter with 16$\times$ as many particles.
BETS also achieves a higher return with this embedding than with a pooled embedding twice as large.

%% file: sections/bg.tex
\section{Background}
\label{sec:background}
BETS plans in POMDPs whose dynamics and observation model are given as input to the problem.
A \textbf{POMDP} is a tuple $\env = (\statespace, \actionspace, \obsspace, \transitionfunc, \obsfunc, \rewardfunc, \discount, \initialstatedist)$ of a set of states $\statespace$, a finite set of actions $\actionspace$, a set of observations $\obsspace$, a transition function $\transitionfunc : \statespace \times \actionspace \rightarrow \dist \statespace$, an observation function $\obsfunc : \statespace \times \actionspace \times \statespace \rightarrow \dist \obsspace$, a reward function $\rewardfunc : \statespace \times \actionspace \times \statespace \rightarrow \reals$, a discount factor $\discount$, and an initial state distribution $\initialstatedist \in \dist \statespace$.
An episode starts from $\statetime{0} \sim \initialstatedist$, and at each step $t$ the agent takes an action $\actiontime{t}$, the environment moves to $\statetime{t+1} \sim \transitionfunc(\cdot \mid \statetime{t}, \actiontime{t})$, and the agent receives the observation $\obstime{t+1} \sim \obsfunc(\cdot \mid \statetime{t}, \actiontime{t}, \statetime{t+1})$ and the reward $\rewardfunc(\statetime{t}, \actiontime{t}, \statetime{t+1})$.
The agent observes its own actions, so what it knows at time $t$ is the \textbf{history} $\history_t \defeq (\actiontime{0}, \obstime{1}, \dots, \actiontime{t-1}, \obstime{t})$, and it acts according to a \textbf{policy} $\jointpolicy : \historyset \to \dist \actionspace$, a mapping from the set $\historyset$ of histories to distributions over actions.
The objective is a policy that maximizes the expected discounted return $\E_{\env, \jointpolicy}[\sum_{t=0}^{\T-1}\discount^{t}\rewardfunc(\statetime{t}, \actiontime{t}, \statetime{t+1})]$ over episodes of $\T$ steps, and the \textbf{action value} $\actionvalue^\jointpolicy(\history_t, \action) \defeq \E_{\env,\jointpolicy}[\returnt \mid \history_t, \actiontime{t} = \action]$ is the expected return from taking $\action$ after $\history_t$ and following $\jointpolicy$ thereafter.

The initial state is random, transitions can be stochastic, and distinct states can emit the same observation, so histories do not identify states; they induce a distribution over $\statespace$.
This distribution is the \textbf{belief} $b_t$, which depends on $\history_t$.
Given $\actiontime{t}$ and $\obstime{t+1}$, Bayes' rule gives the posterior belief,
\nobreak
\begin{equation} \label{eq:bayes}
b_{t+1}(\state^\prime) \propto \sum_{\state \in \statespace} b_t(\state)\, \transitionfunc(\state^\prime \mid \state, \actiontime{t})\, \obsfunc(\obstime{t+1} \mid \state, \actiontime{t}, \state^\prime).
\end{equation}
\nobreak
The update, with an integral in place of the sum when $\statespace$ is continuous, is exact but intractable when the state space is large or continuous.
The belief is a sufficient statistic of the history \citep{kaelbling1998planning}, so an optimal policy and its action values depend on $\history_t$ only through $b_t$.
Planning in a POMDP is therefore planning in a \textbf{belief MDP}, whose states are beliefs, whose transitions apply Eq.~\ref{eq:bayes} to the next observation, and whose values are expected returns under $b_t$.
Its state space is continuous even when $\statespace$ is finite, and the planners of Section~\ref{sec:bg-planning} search with approximations of $b_t$.

\subsection{Approximating Beliefs} \label{sec:bg-beliefs}
\textbf{Particle filters} \citep{gordon1993novel,doucet2009tutorial} approximate $b_t$ by a set of weighted states, or \emph{particles}, $\{\particle, \particleweight\}_{i=1}^n$.
Given $\actiontime{t}$ and $\obstime{t+1}$, the update simulates a successor $\nextparticle \sim \transitionfunc(\cdot \mid \particle, \actiontime{t})$ for every particle and multiplies its weight by the likelihood $\obsfunc(\obstime{t+1} \mid \particle, \actiontime{t}, \nextparticle)$.
When many weights become small, the filter \emph{resamples} by duplicating high-weight particles and discarding the rest.
Resampling never introduces a state the set did not already contain, so repeated resampling concentrates the set on a few distinct states, a failure called \textbf{impoverishment}.
The approximation of $b_t$ then collapses, and the value estimates a planner computes from it inherit the error \citep{poupart2013value}.
Regularization \citep{musso2001improving}, MCMC \citep{gilks2001following}, and locally optimal proposals \citep{doucet2009tutorial} slow the collapse, but every new particle still descends from the surviving set, so loss compounds over the episode.
Preventing weight degeneracy takes exponentially many particles in the dimension of the state \citep{snyder2008obstacles}.

\newcommand{\nbfupdate}{\mc U_{\embedfuncparams,\flowparams}(\bembed,\action,\obs)}

\textbf{Neural Bayesian Filtering} (NBF) \citep{solinas2025neural} approximates the Bayes update of Eq.~\ref{eq:bayes} in the latent space of a \textbf{generative distribution embedding} (GDE) \citep{fishman2025generative}.
A GDE consists of an \textbf{encoder} $\embedfunc : \bigcup_{n \geq 1} \statespace^n \times \R^n \rightarrow \R^{d_{\text{emb}}}$ that maps a weighted set of states to a vector $\bembed$, and a conditional \textbf{generator} $\generator(\bembed)$ that maps the vector back to sample states.
The encoder is \emph{distributionally invariant} \citep{fishman2025generative}: it depends on a weighted set only through the set's weighted empirical measure, so reordering the set or duplicating every sample leaves its output unchanged.
The NBF \textbf{update} $\bembed^\prime \leftarrow \nbfupdate$ generates $n$ particles from $\generator(\bembed)$, propagates each through $\transitionfunc$, weights it by $\obsfunc$, and embeds the weighted set with $\embedfunc$.
Each update regenerates its particle set from the embedding, so the collapse that resampling causes in a particle filter does not accumulate across steps.
If the particle set loses most or all of the belief's support in a single update, the approximation can collapse irrecoverably.
In practice, this happens only in extreme cases, such as a sharp likelihood on a belief far from those BETS visits during training.

\subsection{Planning with Partial Observability} \label{sec:bg-planning}
Full-width planners expand every branch with exact belief updates \citep{ross2008online}, which limits them to small state and observation spaces.
Monte Carlo planners \citep{kocsis2006bandit,browne2012survey} scale better by simulating states drawn from an approximate belief through the environment dynamics.
POMCP \citep{silver2010monte} runs UCT over histories with a particle set at each node, and POMCPOW \citep{sunberg2018online} adds progressive widening for large observation spaces.
DESPOT \citep{somani2013despot} and AdaOPS \citep{wu2021adaptive} search a sparse tree over a fixed set of sampled \emph{scenarios} and prune it with upper and lower bounds on the value.
These planners have no learned components, so their search is only as informed as their rollout policies and bounds, but they need no training phase.
BetaZero \citep{moss2023betazero} approximates each node's belief with a particle filter, and uses learned policy and value networks to guide search.
These networks condition on the mean and standard deviation of features of the particle set and are trained on targets from BetaZero's own searches \citep{silver2018general}.

Summary statistics of particle features can discard the structure a decision depends on.
Consider two cells, each of which may hold a hidden object, so the state is $x \in \{0,1\}^2$ with $x_j = 1$ when cell $j$ is occupied.
The belief $b_{\mathrm{excl}}$ is uniform over $\{01, 10\}$, where exactly one cell is occupied, and $b_{\mathrm{co}}$ is uniform over $\{00, 11\}$, where both or neither are.
Both beliefs give each cell probability $\tfrac12$ of holding the object, so their per-cell means and standard deviations are identical, yet their supports are disjoint.
An action rewarded only when both cells are occupied has zero expected reward under $b_{\mathrm{excl}}$ and positive expected reward under $b_{\mathrm{co}}$.
A network that conditions on per-variable statistics must assign the two beliefs the same policy and value.
A generator conditioned on per-variable statistics fails in the same way, since it must produce one distribution for both beliefs and so places at least half of its mass on configurations that one of them rules out.
A planner that simulates those configurations estimates values on states the belief rules out.
The next section builds a planner whose networks and generator condition on an embedding that represents this dependence explicitly.

%% file: sections/bets.tex
\section{Belief Embedding Tree Search}
\label{sec:bets}

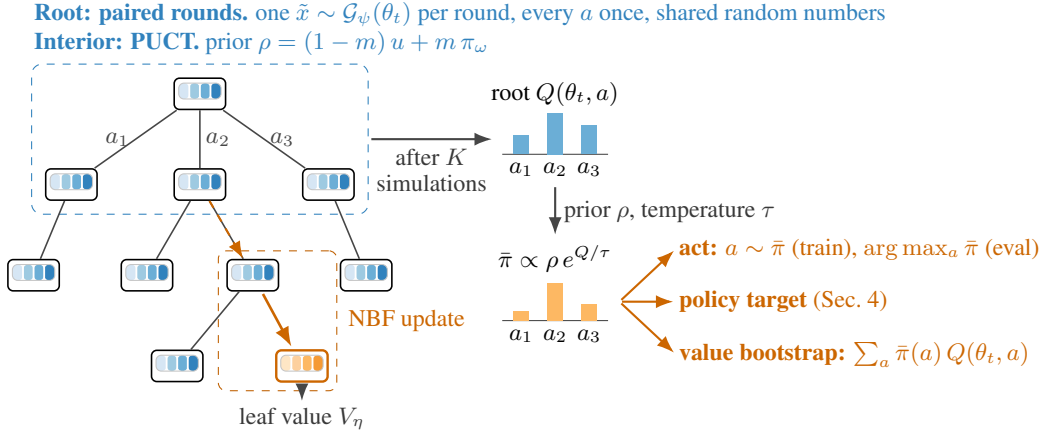
\begin{figure}[t]
    \centering
    \input{fig/bets-search.tikz}
    \caption{\textbf{BETS.}
    Each round starts every root action from the same sampled state and interior actions are selected via PUCT. Expansion costs one NBF update.}
    \label{fig:bets}
\end{figure}

Belief Embedding Tree Search (BETS) is a decision-time planner for POMDPs that searches over a tree of embeddings (Figure~\ref{fig:bets}).
At each step, BETS roots the tree at the root's embedding and returns a policy over actions that determines the agent's next action.
Every node in the tree stores one embedding $\bembed$, and expanding a leaf is exactly one NBF update.
A policy network $\policynetdef$ and a value network $\valuenetdef$ guide the search.
Both networks condition on the embedding and a small vector $\obsfeat$ of features of the observed part of the state, such as the fraction of the horizon elapsed (Appendix~\ref{app:domains}).
Conditioning on $\bembed$ rather than on per-variable statistics lets the networks assign different policies and values to beliefs that share those statistics.

\subsection{Search}
\label{sec:bets-search}
At decision time, BETS follows the standard Monte Carlo tree search procedure of selection, expansion, evaluation, and backup \citep{kocsis2006bandit,browne2012survey}.
It runs $K$ simulations of this procedure through a tree rooted at the belief embedding $\bembed_t$.

\paragraph{Root.}
BETS groups the $K$ simulations into $K / |\actionspace|$ rounds, and each round selects every root action exactly once.
At the start of a round, BETS samples a state $\tilde{\state} \sim \generator(\bembed_t)$, and every action in the round starts from $\tilde{\state}$ with the same random numbers, so the actions' returns are \emph{paired} \citep{veness2011variance}.
The decision depends only on the differences between root action values, and pairing reduces the variance that the sampled state adds to those differences.
When UCT \citep{kocsis2006bandit} or PUCT \citep{rosin2011multi,silver2017mastering} selects the root action, each action is evaluated on its own sampled states, and that state sampling variance can influence the relative value estimates of actions for finite $K$.
Selecting every action once per round leaves root visit counts uniform, so the decision and the policy target are computed from the root action values instead (Section~\ref{sec:bets-root}).

\paragraph{Interior.}
Below the root, the traversal applies the selected action to $\tilde{\state}$, samples the next state and observation from $\transitionfunc$ and $\obsfunc$, and descends to the child node for that action and observation if one exists.
At an interior node with embedding $\bembed$, the traversal selects the action maximizing PUCT,
\[
\operatorname*{arg\,max}_{\action \in \actionspace} \; \bar{\actionvalue}(\bembed, \action) + c \, \policyprior(\action \mid \bembed, \obsfeat) \, \frac{\sqrt{N(\bembed)}}{1 + N(\bembed, \action)},
\]
where $N(\bembed)$ and $N(\bembed, \action)$ are the node and edge visit counts, $\bar{\actionvalue}$ is the action value estimate normalized to $[-1, 1]$ within the node \citep{schrittwieser2020mastering}, and $c$ is the exploration constant.
The prior $\policyprior$ mixes the policy network with the uniform distribution $\unif$ over legal actions,
\[
\policyprior(\action \mid \bembed, \obsfeat) \defeq (1 - \priormix)\, \unif(\action) + \priormix\, \policynet(\action \mid \bembed, \obsfeat),
\]
so no action's prior reaches zero and BETS keeps exploring regardless of policy network bias.

\paragraph{Expansion, evaluation, and backup.}
When the current node has no child for the action $\action$ and observation $\obs$ of the traversal's last step, BETS expands by adding a new leaf.
If the current node lies $d$ levels below the root with embedding $\bembed_{t+d}$, the leaf's embedding is computed with an NBF update, $\bembed_{t+d+1} \leftarrow \mc U_{\embedfuncparams,\flowparams}(\bembed_{t+d}, \action, \obs)$.
The value network evaluates the leaf, $\valuenet(\bembed_{t+d+1}, \obsfeat_{t+d+1})$, with no rollout.
The value backed up through the traversal is the discounted sum of the rewards generated below it, plus the leaf value.
The visit counts along the path are then incremented.
Each simulation applies the NBF update, the policy network, and the value network once, and every node stores one embedding in place of a particle set.

\subsection{Action Selection}
\label{sec:bets-root}
After $K$ simulations, the root node holds a paired estimate $\actionvalue(\bembed_t, \action)$ for every action.
BETS turns these estimates into a distribution over actions with one step of KL-regularized policy improvement \citep{abdolmaleki2018maximum},
\[
\pibar \defeq \operatorname*{arg\,max}_{\pi^\prime \in \dist \actionspace} \; \sum_\action \pi^\prime(\action)\, \actionvalue(\bembed_t, \action) - \readtemp\, \kl{\pi^\prime}{\policyprior(\cdot \mid \bembed_t, \obsfeat_t)},
\]
where the temperature $\readtemp$ sets how far the search can move the policy away from the prior.
The maximizer has the closed form $\pibar(\action) \propto \policyprior(\action \mid \bembed_t, \obsfeat_t) \exp(\actionvalue(\bembed_t, \action) / \readtemp)$, which is one mirror descent step from the prior with step size $1/\readtemp$ \citep{beck2003mirror}.
AlphaZero's visit-count distribution approximates the solution of this problem with the KL taken in the other direction \citep{grill2020monte}, and \citet{danihelka2022policy} use this closed form.
BETS evaluates this closed form exactly.

During training, the agent acts by sampling $\action \sim \pibar$; at evaluation, it takes $\arg\max_\action \pibar(\action)$.
The same $\pibar$ is the policy network's training target, and its expected root value $\sum_\action \pibar(\action)\, \actionvalue(\bembed_t, \action)$ is the bootstrap for the value network's target (Section~\ref{sec:targets}).
After BETS acts, one more NBF update computes the next root embedding from the action taken and the observation received.

\subsection{Variable-Coupling Embeddings}
\label{sec:vc}
Standard GDE encoders compute a feature vector for each sample and pool the per-sample features over the set \citep{fishman2025generative}.
Each per-sample feature depends on every variable of its sample, so in principle the pooled features can capture how the variables of a sample depend on one another.
The two beliefs in the example of Section~\ref{sec:bg-planning} give each cell the same probability of holding the object, so the dependence between the two cells is all that separates them.
In practice, the encoder has to discover the dependence inside generic per-sample features, and if the dependence does not survive the pooling, the two beliefs receive the same embedding.
We extend GDEs by giving the encoder and the generator an explicit representation of the co-occurrence between pairs of variables.

\begin{figure}[t]
  \centering
  \scalebox{0.9}{\input{fig/variable-coupling.tikz}}
  \caption{\textbf{The variable-coupling stream.}
  Pooling over samples gives one summary per variable and a variable-by-variable co-occurrence matrix.
  Attention over the variable summaries, biased by the co-occurrence matrix, produces coupling summaries, which are concatenated with $\mu$ to form $\bembed$.}
  \label{fig:varcoupling}
\end{figure}
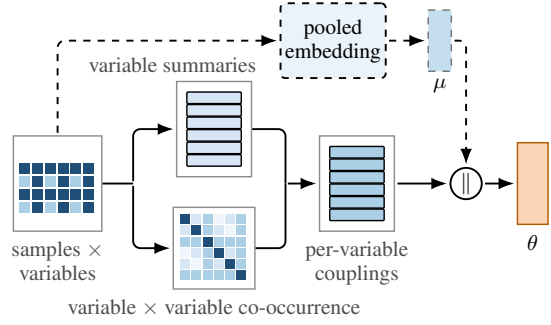

A discrete state is a vector $x \in \{0, \dots, q-1\}^D$ of $D$ variables.
The \emph{pooled stream} is a set transformer \citep{lee2019set} whose per-sample features are averaged over the set to give the pooled embedding $\mu$.
A second stream, which we call the \textbf{variable-coupling stream}, averages a joint feature of each pair of variables over the same samples.
From these pairwise averages it forms a coupling summary for each variable that describes how the variable relates to the others (Figure~\ref{fig:varcoupling}).
The $D$ coupling summaries are combined with $\mu$ to form $\bembed$.
The policy and value networks take all of $\bembed$ as input, so they condition on every coupling summary as well as on $\mu$.
Both streams average over samples, so $\bembed$ remains distributionally invariant, and both are trained jointly with the generator.

BETS pairs the variable-coupling stream with an autoregressive generator that treats each variable as a token and factorizes the belief over them, $p_\flowparams(x \mid \bembed) = \prod_{j=1}^{D} p_\flowparams(x_j \mid x_{<j}, \bembed)$, where $p_\flowparams(\cdot \mid \bembed)$ is the density of $\generator(\bembed)$.
The generator conditions factor $j$ on the preceding variables $x_{<j}$, the shared pooled embedding $\mu$, and the coupling summary of variable $j$ (Appendix~\ref{app:ar-conditioning}).
Each factor is a normalized categorical, so the product is an exact joint likelihood computable in one forward pass.

%% file: fig/bets-search.tikz
\begin{tikzpicture}[every node/.style={font=\small}, oplabel/.append style={font=\footnotesize}]

\node[treenode] (R)  at (2.6, 0.0) {\bstrip[1.4mm]};
\node[treenode] (A1) at (0.9,-1.2) {\bstrip[1.4mm]};
\node[treenode] (A2) at (2.6,-1.2) {\bstrip[1.4mm]};
\node[treenode] (A3) at (4.3,-1.2) {\bstrip[1.4mm]};
\draw[treeedge] (R)--(A1) node[midway, left=1pt, oplabel] {$\action_1$};
\draw[treeedge] (R)--(A2) node[midway, right=1pt, oplabel] {$\action_2$};
\draw[treeedge] (R)--(A3) node[midway, right=1pt, oplabel] {$\action_3$};

\node[treenode] (B1) at (0.4,-2.4) {\bstrip[1.4mm]};
\node[treenode] (B2) at (1.9,-2.4) {\bstrip[1.4mm]};
\node[treenode] (B3) at (3.3,-2.4) {\bstrip[1.4mm]};
\node[treenode] (B4) at (4.8,-2.4) {\bstrip[1.4mm]};
\draw[treeedge] (A1)--(B1);
\draw[treeedge] (A2)--(B2);
\draw[treeedge] (A2)--(B3);
\draw[treeedge] (A3)--(B4);
\node[treenode] (C1) at (2.3,-3.6) {\bstrip[1.4mm]};
\draw[treeedge] (B3)--(C1);
\node[treenew]  (L)  at (3.95,-3.6) {\ostrip[1.4mm]};

\draw[seledge] (A2)--(B3);
\draw[arr, orange!80!black, line width=1pt] (B3)--(L);
\begin{scope}[on background layer]
  \node[draw=orange!75!black, dashed, rounded corners=3pt, inner sep=3pt, fit=(B3)(L)] (nbfbox) {};
\end{scope}
\node[oplabel, text=orange!80!black, anchor=west, align=left] at ([xshift=3pt]nbfbox.east) {NBF update};
\draw[arr, dashed, gray!55!black] (L) -- ++(0,-0.5) node[below, oplabel, text=gray!45!black] {leaf value $\valuenet$};

\begin{scope}[on background layer]
  \node[draw=mplBluesD, rounded corners=4pt, dashed, inner sep=5pt, fit=(R)(A1)(A3)] (rootbox) {};
\end{scope}
\node[oplabel, text=mplBluesD, anchor=south west, align=left] at ([yshift=2pt]rootbox.north west)
  {\textbf{Root: paired rounds.} one $\tilde\state \sim \generator(\bembed_t)$ per round, every $\action$ once, shared random numbers\\[1pt]
   \textbf{Interior: PUCT.} prior $\policyprior = (1-\priormix)\,\unif + \priormix\,\policynet$};

\begin{scope}[shift={(6.6,-0.8)}]
  \draw[gray!55!black] (0,0)--(1.4,0);
  \fill[mplBluesC] (0.15,0) rectangle (0.35,0.25);
  \fill[mplBluesC] (0.60,0) rectangle (0.80,0.55);
  \fill[mplBluesC] (1.05,0) rectangle (1.25,0.38);
  \node[oplabel] at (0.25,-0.2) {$\action_1$};
  \node[oplabel] at (0.70,-0.2) {$\action_2$};
  \node[oplabel] at (1.15,-0.2) {$\action_3$};
  \node[oplabel, anchor=south] at (0.7,0.6) {root $\actionvalue(\bembed_t,\action)$};
  \coordinate (qwest) at (-0.1,0.25);
  \coordinate (qsouth) at (0.7,-0.4);
\end{scope}
\begin{scope}[shift={(6.6,-3.0)}]
  \draw[gray!55!black] (0,0)--(1.4,0);
  \fill[mplOrgC] (0.15,0) rectangle (0.35,0.12);
  \fill[mplOrgC] (0.60,0) rectangle (0.80,0.50);
  \fill[mplOrgC] (1.05,0) rectangle (1.25,0.22);
  \node[oplabel] at (0.25,-0.2) {$\action_1$};
  \node[oplabel] at (0.70,-0.2) {$\action_2$};
  \node[oplabel] at (1.15,-0.2) {$\action_3$};
  \node[oplabel, anchor=south] at (0.7,0.6) {$\pibar \propto \policyprior\, e^{\actionvalue/\readtemp}$};
  \coordinate (pinorth) at (0.7,1.1);
  \coordinate (pieast) at (1.55,0.25);
\end{scope}
\draw[arr, gray!55!black] (rootbox.east) -- (rootbox.east -| qwest)
  node[midway, below=2pt, oplabel, text=gray!45!black] {after $K$\\simulations};
\draw[arr, gray!55!black] (qsouth) -- (pinorth)
  node[midway, right=2pt, oplabel, text=gray!45!black] {prior $\policyprior$, temperature $\readtemp$};
\node[oplabel, text=orange!80!black, anchor=west, align=left] (out1) at (8.9,-2.05)
  {\textbf{act:} $\action \sim \pibar$ (train), $\arg\max_\action \pibar$ (eval)};
\node[oplabel, text=orange!80!black, anchor=west, align=left] (out2) at (8.9,-2.75)
  {\textbf{policy target} (Sec.~\ref{sec:training})};
\node[oplabel, text=orange!80!black, anchor=west, align=left] (out3) at (8.9,-3.45)
  {\textbf{value bootstrap:} $\sum_\action \pibar(\action)\,\actionvalue(\bembed_t,\action)$};
\draw[arr, orange!80!black] (pieast) -- (out1.west);
\draw[arr, orange!80!black] (pieast) -- (out2.west);
\draw[arr, orange!80!black] (pieast) -- (out3.west);

\end{tikzpicture}

%% file: fig/variable-coupling.tikz
\begin{tikzpicture}[
    >=Latex,
    every node/.style={font=\small},
    /utils/exec=\definecolor{mplBluesA}{HTML}{F0F6FC}%
        \definecolor{mplBluesB}{HTML}{D6E6F4}%
        \definecolor{mplBluesC}{HTML}{ABD0E6}%
        \definecolor{mplBelief}{HTML}{1F4E79}%
        \definecolor{mplOrange}{HTML}{F4A261},
    block/.style={
        draw, rounded corners=2pt, thick, fill=mplBluesB,
        minimum width=14mm, minimum height=14mm,
        align=center, inner sep=2pt, font=\footnotesize,
    },
    existing/.style={block, dashed, fill=mplBluesB!50},
    data/.style={
        draw=black!45, semithick, sharp corners, fill=white,
        minimum width=14mm, minimum height=14mm,
        align=center, inner sep=2pt,
    },
    gridp/.style={data, minimum width=13mm},
    matp/.style={data, minimum width=12mm, minimum height=12mm},
    tokp/.style={data, minimum width=11mm},
    op/.style={
        draw, thick, circle, fill=white,
        inner sep=0.5pt, minimum size=4.6mm, font=\footnotesize,
    },
    tokenbar/.style={
        draw, semithick, rounded corners=0.5pt, fill=mplBluesC,
        minimum width=8mm, minimum height=1.7mm, inner sep=0pt,
    },
    tokenbarl/.style={tokenbar, fill=mplBluesB},
    summaryd/.style={
        draw=black!55, semithick, sharp corners, dashed, fill=mplBluesC!70,
        minimum width=3.4mm, minimum height=9mm, inner sep=1pt,
    },
    final/.style={
        draw=orange!70!black, semithick, sharp corners, fill=mplOrange!45,
        minimum width=4.5mm, minimum height=12mm, inner sep=1pt,
    },
    lbl/.style={font=\small, inner sep=1pt, align=center},
    sub/.style={font=\footnotesize, inner sep=1pt, align=center, text=black!70},
    arr/.style={->, thick, shorten >=1pt, shorten <=1pt},
    arrd/.style={arr, dashed},
    rail/.style={arr, rounded corners=3pt},
    raild/.style={rail, dashed},
    merge/.style={thick, rounded corners=3pt},
]

\node[gridp] (G) at (0,0) {};
\begin{scope}[shift={(G.center)}]
    \def\p{1.9mm}
    \foreach \r in {0,1,2,3} {
        \foreach \c in {0,1,2,3,4,5} {
            \pgfmathtruncatemacro{\parity}{mod(\r+2*\c+\c*\r,2)}
            \ifnum\parity=0 \def\cc{mplBelief}\else\def\cc{mplBluesC}\fi
            \fill[\cc] ($(-5.7mm,3.0mm) + (\c*\p,-\r*\p)$) rectangle ++(1.6mm,-1.6mm);
        }
    }
\end{scope}
\node[below=1mm of G, sub] {samples $\times$\\variables};

\coordinate (F) at ($(G.east)+(4.5mm,0)$);

\node[tokp] (vtok) at (23mm, 8mm) {};
\begin{scope}[shift={(vtok.center)}]
    \foreach \i in {0,1,2,3,4,5} {
        \node[tokenbarl] at (0, {0.46 - 0.184*\i}) {};
    }
\end{scope}
\node[sub, anchor=south east] at ($(vtok.north east)+(1mm,0.8mm)$) {variable summaries};

\node[matp] (C) at (23mm, -9.5mm) {};
\begin{scope}[shift={(C.center)}]
    \def\p{1.65mm}
    \foreach \r in {0,1,2,3,4,5} {
        \foreach \c in {0,1,2,3,4,5} {
            \ifnum\r=\c
                \def\cc{mplBelief}
            \else
                \pgfmathtruncatemacro{\s}{mod(\r*\c+\r+\c,3)}
                \ifcase\s \def\cc{mplBluesA}\or \def\cc{mplBluesB}\else \def\cc{mplBluesC}\fi
            \fi
            \fill[\cc] ($(-4.95mm,4.95mm) + (\c*\p,-\r*\p)$) rectangle ++(1.4mm,-1.4mm);
        }
    }
\end{scope}
\node[below=1mm of C, sub] {variable $\times$ variable co-occurrence};

\coordinate (J) at (33.5mm, 0);

\node[tokp] (tok) at (44mm, 0) {};
\begin{scope}[shift={(tok.center)}]
    \foreach \i in {0,1,2,3,4,5} {
        \node[tokenbar] at (0, {0.46 - 0.184*\i}) {};
    }
\end{scope}
\node[below=1mm of tok, sub] {per-variable\\couplings};

\node[existing, minimum height=11mm] (ppool) at (40.5mm, 21mm) {pooled\\embedding};
\node[summaryd, right=6mm of ppool] (mu) {};
\node[below=0.5mm of mu, lbl] {$\mu$};

\node[op, right=8mm of tok] (cat) {$\|$};
\node[final, right=5mm of cat] (z) {};
\node[below=1mm of z, lbl] {$\bembed$};

\draw[thick] (G) -- (F);
\draw[rail] (F) |- (vtok.west);
\draw[rail] (F) |- (C.west);

\draw[merge] (vtok.east) -| (J);
\draw[merge] (C.east) -| (J);
\draw[arr] (J) -- (tok.west);

\draw[arr] (tok) -- (cat);
\draw[arr] (cat) -- (z);

\draw[raild] (G.north) |- (ppool.west);
\draw[arrd]  (ppool) -- (mu);
\draw[raild] (mu.east) -| (cat.north);

\end{tikzpicture}

%% file: sections/training.tex
\section{Training} \label{sec:training}
BETS learns three models: the belief model, which is the GDE encoder and generator that NBF uses in its update, and the policy and value networks that guide the search.
The beliefs BETS visits depend on its policy, so the beliefs the model must represent change as the networks train and the search policy improves.
All three models train together on episodes that BETS collects.

Training alternates between gradient steps and collection rounds.
In each \emph{collection round}, a set of collectors runs BETS with the parameters from the most recent gradient step, and each collector takes a fixed number of environment steps.
The consecutive transitions that one collector records in one round form a \emph{fragment}, and an episode that ends inside a fragment is reset, so the fragment continues with the next episode.
Before the first gradient step, a fixed number of collection rounds run with the initial model parameters.
Collection rounds repeat at a fixed interval of gradient steps, and every gradient step draws one batch for each of the three losses from the fragments of the most recent rounds.
The policy and value networks are trained on the embeddings $\bembed_t$ stored during collection, so their losses never reach the encoder and the GDE is trained on the belief loss alone.
The exploration constant $c$, the mix weight $\priormix$, and the temperature $\readtemp$ are annealed from exploratory to greedy values over the first part of training (Appendix~\ref{app:hyperparameters}).

\subsection{Belief Model} \label{sec:belief-training}
Each training example for the GDE is a tuple $(\bembed_t, \actiontime{t}, \obstime{t+1}, \statetime{t+1})$ stored from a collected episode.
For each example, the NBF update generates particles from $\generator(\bembed_t)$, propagates them under $\actiontime{t}$, and weights them by the likelihood of $\obstime{t+1}$, which yields the weighted set $\weightedparticleset_{t+1}$ and its embedding $\bembed_{t+1} = \embedfunc(\weightedparticleset_{t+1})$.
The update runs at training time, so the particle sets the belief model embeds are regenerated from the generator being trained rather than loaded from storage.
At the start of an episode, the particles are drawn from $\initialstatedist$ instead.

The target is the hidden state $\statetime{t+1}$ that the environment realized after $\actiontime{t}$.
Though the agent never observes $\statetime{t+1}$, it is the ground truth state for simulation, so it is available to compute the loss at training time.
The belief model minimizes the realized-state objective of \citet{solinas2025neural},
\begin{equation} \label{eq:realized-state-loss}
\shortloss \defeq \E\big[\ell_\flowparams\big(\statetime{t+1};\, \embedfunc(\weightedparticleset_{t+1})\big)\big],
\end{equation}
where the expectation is over the collected transitions and over the particle sets the NBF update generates from them.
$\ell_\flowparams(\state; \bembed) = -\log p_\flowparams(\state \mid \bembed)$ is the GDE's negative log-likelihood of the single target $\state$ given the embedding $\bembed = \embedfunc(\weightedparticleset_{t+1})$.\footnote{On LightDark, $\ell_\flowparams$ is instead the conditional flow-matching regression at $\statetime{t+1}$ (Appendix~\ref{app:fm}).}
Gradients pass through the embedding of the set and through the generator, but not through the sampling that produced the set.

BETS's actions depend on the hidden state only through the history, so each $\statetime{t+1}$ is a sample from the true posterior $b_{t+1}$ given the history $\history_{t+1}$.
The particle set $\weightedparticleset_{t+1}$ is computed from that history and the filter's own random draws, and the \textbf{set posterior} $p(\cdot \mid \weightedparticleset_{t+1})$ is the distribution of $\statetime{t+1}$ given the set alone.
It equals $b_{t+1}$ when the set retains all the information that the history carries about the state, and in general it is the belief with the lowest expected log loss among all beliefs computed from the set.
By the decomposition of \citet[Proposition~4.1]{solinas2025neural}, minimizing $\shortloss$ for fixed sets is equivalent to minimizing $\E\big[\mathrm{KL}\big(p(\cdot \mid \weightedparticleset_{t+1}) \,\|\, p_\flowparams(\cdot \mid \bembed_{t+1})\big)\big]$, the expected KL divergence from the set posterior to the GDE's belief $p_\flowparams(\cdot \mid \bembed_{t+1})$.

\subsection{Policy and Value Targets} \label{sec:targets}
The policy and value networks are trained on targets derived from the search \citep{anthony2017thinking,silver2017mastering}.
At step $t$, the root gives the improved policy $\pibar_t$ and its value $\vbar_t \defeq \sum_\action \pibar_t(\action)\, \actionvalue(\bembed_t, \action)$, and the transition $(\bembed_t, \obsfeat_t, \actiontime{t}, r_t, \pibar_t, \vbar_t)$ is stored, where $r_t$ is the reward of step $t$.
The \textbf{policy target} is $\pibar_t$, and the policy network minimizes the cross-entropy $\mathrm{H}(\pibar_t, \policynet(\cdot \mid \bembed_t, \obsfeat_t))$.

The \textbf{value target} is the $\lambda$-return bootstrapped from the root values $\vbar$ \citep{sutton1998reinforcement},
\[
z^\lambda_t \defeq r_t + \discount\big[(1 - \lambda)\, \vbar_{t+1} + \lambda\, z^\lambda_{t+1}\big],
\]
computed backward over each fragment, with $z^\lambda_{t+1} \defeq \vbar_{t+1}$ past the end of a fragment and $\vbar_{t+1} = z^\lambda_{t+1} = 0$ at terminal states.
The value network is trained as a classifier over a fixed grid of bins.
Each $\lambda$-return is squashed by $f(u) = \operatorname{sign}(u)\big(\sqrt{|u| + 1} - 1\big) + \epsilon u$ \citep{pohlen2018observe}, and $f(z^\lambda_t)$ is encoded as a two-hot distribution over the bins \citep{schrittwieser2020mastering}.
Training minimizes the cross-entropy between the value network's prediction and this two-hot target.
During search, a leaf's value is $f^{-1}$ applied to the mean of the predicted distribution over the bins.

%% file: sections/experiments.tex
\section{Experiments} \label{sec:experiments}
We evaluate BETS on six configurations of three POMDP domains: RockSample(15,15) and RockSample(20,20) \citep{smith2004heuristic}, LightDark(5) and LightDark(10) \citep{platt2010belief}, and PocMan on a medium maze (M) and the large maze (L) of \citet{silver2010monte} (Appendix~\ref{app:domains}).
The domains differ in belief structure.
In RockSample, the agent's position is known, and the belief is a product of 15 or 20 independent Bernoulli distributions, one per rock, over $2^{15}$ or $2^{20}$ joint configurations.
In LightDark, the agent moves along a line, and the belief is a one-dimensional density over its position that stays broad in the dark and sharpens near the light.
In PocMan, it is a joint distribution over the positions and headings of three or four ghosts, which move through the same maze, chasing or running away from the agent when it is nearby.

\paragraph{Baselines.}
We compare against \textbf{BetaZero} \citep{moss2023betazero} and three planners without learned components: \textbf{AdaOPS} \citep{wu2021adaptive}, \textbf{DESPOT} \citep{somani2013despot}, and \textbf{POMCPOW} \citep{sunberg2018online}.
At evaluation, all four use the same number of simulations per decision and particles as BETS, and otherwise use the settings of \citet{moss2023betazero} on RockSample and LightDark (Appendix~\ref{app:hyperparameters}).

\paragraph{Protocol.}
BETS is evaluated in two ways on the same model: with \textbf{search}, acting by $\arg\max_\action \pibar(\action)$ with the search settings training ends with, and with the \textbf{policy} network alone, acting by $\arg\max_\action \policynet(\action \mid \bembed, \obsfeat)$.
We tuned the training hyperparameters of BETS on every domain and those of BetaZero on PocMan, where \citet{moss2023betazero} give no settings.
BETS and BetaZero are trained at three seeds, and each trained model is evaluated on 300 episodes at each of three evaluation seeds.
AdaOPS, DESPOT, and POMCPOW have no learned components, so each runs three evaluation seeds with 300 episodes each.

\begin{table}[t]
\caption{Mean discounted return with the SE across training seeds (BETS, BetaZero) or evaluation seeds (AdaOPS, DESPOT, POMCPOW).
AdaOPS has no bounds implementation for PocMan.}
\label{tab:results}
\centering
\footnotesize
\setlength{\tabcolsep}{2.9pt}
\begin{threeparttable}
\begin{tabular}{@{}lcccccc@{}}
\toprule
 & \multicolumn{2}{c}{BETS} & & & & \\
\cmidrule(lr){2-3}
Domain & search & policy & BetaZero & AdaOPS & DESPOT & POMCPOW \\
\midrule
RockSample(15,15) & 20.78 $\pm$ 0.09 & 20.65 $\pm$ 0.10 & 19.79 $\pm$ 0.18 & \textbf{21.45 $\pm$ 0.10} & 19.45 $\pm$ 0.10 & 11.45 $\pm$ 0.08 \\
RockSample(20,20) & \textbf{14.97 $\pm$ 0.13} & 13.91 $\pm$ 0.16 & 11.30 $\pm$ 0.15 & 10.77 $\pm$ 0.20$^\dagger$ & 0.28 $\pm$ 0.02$^\dagger$ & 4.33 $\pm$ 0.18$^\dagger$ \\
LightDark(5) & 3.61 $\pm$ 0.10 & 3.77 $\pm$ 0.12 & 3.76 $\pm$ 0.15 & 3.08 $\pm$ 0.03$^\dagger$ & 2.29 $\pm$ 0.09 & 2.82 $\pm$ 0.25 \\
LightDark(10) & 13.01 $\pm$ 0.41 & 13.88 $\pm$ 0.28 & \textbf{14.43 $\pm$ 0.33} & 3.18 $\pm$ 0.40$^\dagger$ & 0.33 $\pm$ 0.05 & 0.98 $\pm$ 0.13 \\
PocMan(M) & \textbf{78.08 $\pm$ 0.54} & 72.42 $\pm$ 2.05 & 40.87 $\pm$ 3.26 & -- & 54.20 $\pm$ 0.34 & 36.34 $\pm$ 1.93 \\
PocMan(L) & \textbf{78.95 $\pm$ 0.23} & 77.47 $\pm$ 0.46 & 69.87 $\pm$ 1.64 & -- & 60.49 $\pm$ 0.71 & 39.91 $\pm$ 1.41 \\
\bottomrule
\end{tabular}
\begin{tablenotes}
\item[$^\dagger$] Without domain heuristics: exit-policy lower and fixed upper bounds for AdaOPS and DESPOT on RockSample(20,20), fixed bounds for AdaOPS on LightDark, and no rollout heuristic for POMCPOW.
\end{tablenotes}
\end{threeparttable}
\end{table}

\subsection{Results} \label{sec:exp-return}
Table~\ref{tab:results} shows that BETS earns the highest return on RockSample(20,20) and on both PocMan mazes, where baseline search heuristics or belief approximations fail.
It trails AdaOPS on RockSample(15,15), where AdaOPS's heuristic bounds are feasible, and BetaZero on LightDark(10).
On RockSample(20,20) computing the heuristic upper bounds of AdaOPS and DESPOT is infeasible, so both run on a fixed upper bound, which reduces discounted return.

BetaZero also learns to guide its search, so we attribute most of the difference from BetaZero to what its networks condition on: the mean and standard deviation of each state variable over its particle set.
Either the particle set or its summary statistics can make a poor input to the policy and value networks.
On RockSample(20,20) the belief is a product of Bernoulli distributions, so the per-rock means are a sufficient statistic for it.
We attribute the gap there to BetaZero's particle set, which covers a small fraction of the $2^{20}$ configurations.
Each resampling step of the bootstrap filter duplicates the surviving particles, and with deterministic transitions the duplicates stay identical, so the per-rock means rest on ever fewer distinct configurations that carry disproportionate weight, whereas BETS regenerates its particles from the embedding at every update.
A resampler that redraws each rock from its per-rock mean would restore distinct configurations, but it relies on the factorization, which is a domain-specific property.

On PocMan we attribute the gap to the summary statistics, which lose two kinds of structure.
A ghost that may be in any of several corridors has a multimodal belief, and the mean and standard deviation of its position cannot distinguish those corridors from the region between them.
These statistics also discard the dependence between state variables, such as ghost positions and headings, which is what the variable-coupling stream was designed to model.

BetaZero matches BETS on LightDark(5) and leads on LightDark(10).
With a one-dimensional state the particle set approximates the belief well.
Far from the light, the Gaussian likelihood spreads the weight over many particles even under an inaccurate proposal, and near it the belief itself narrows almost to a point, so its mean and standard deviation are close to a sufficient statistic for it.
The policy network alone matches search on LightDark(5) and exceeds it on LightDark(10).
We hypothesize that, because every sampled observation is new, the tree stays shallow and a single search compares actions mostly through the value network's estimates at a few sampled leaves, whose noise the policy network averages out by training on $\pibar$ from many searches.
On RockSample(20,20) and both PocMan mazes, the policy network alone earns more return than every baseline does with search.

\paragraph{Effect of Variable Coupling on PocMan.}
\textbf{VC} is the BETS configuration of Table~\ref{tab:results}, and \textbf{Pooled} removes its variable-coupling stream and doubles the pooled embedding to 128 dimensions, which more than doubles the model's capacity.
Everything else is held fixed, and both are trained with three seeds on PocMan(M).
Table~\ref{tab:vc} shows that Pooled earns less return than VC with both search and the policy network, so the added capacity does not replace the variable-coupling stream.

Pooled has to discover the dependence between ghost positions and headings inside generic per-sample features.
With search, Pooled also clears the maze in fewer than half as many episodes, and removing the stream costs more discounted return.
Search scores its leaves with the value network and simulates states drawn from the belief, so the larger difference under search suggests that the coupling summaries improve both the value estimates and the simulated states.

\begin{table}[t]
\captionsetup[subtable]{justification=centering}
\caption{Variable coupling and value prediction on PocMan(M).}
\label{tab:pocman}
\centering
\footnotesize
\setlength{\tabcolsep}{4pt}
\begin{minipage}[c]{0.53\textwidth}
\centering
\begin{tabular}{@{}lcc@{}}
\toprule
& VC & Pooled \\
\midrule
Discounted return (policy) & \textbf{72.42 $\pm$ 2.05} & 65.39 $\pm$ 0.72 \\
Discounted return (search) & \textbf{78.08 $\pm$ 0.54} & 68.20 $\pm$ 0.48 \\
Maze clears (search) & \textbf{14.1\%} & 6.7\% \\
\bottomrule
\end{tabular}
\end{minipage}\hfill
\begin{minipage}[c]{0.45\textwidth}
\centering
\setlength{\tabcolsep}{2.5pt}
\begin{tabular}{@{}lcccc@{}}
\toprule
Episodes Collected & 5000 & 10000 & 25000 & 50000 \\
\midrule
VC & \textbf{0.256} & \textbf{0.280} & \textbf{0.315} & \textbf{0.341} \\
Pooled & 0.245 & 0.276 & 0.311 & 0.330 \\
Moments & 0.218 & 0.250 & 0.285 & 0.311 \\
\bottomrule
\end{tabular}
\end{minipage}

\begin{subtable}[t]{0.53\textwidth}
\caption{Mean discounted return over three training seeds.}
\label{tab:vc}
\end{subtable}\hfill
\begin{subtable}[t]{0.45\textwidth}
\caption{Held-out $R^2$ of value networks.}
\label{tab:value-rep}
\end{subtable}
\end{table}

\paragraph{Value Prediction from Belief Representations.}
We collect 50000 PocMan(M) episodes with the policy network from one training seed, while a bootstrap particle filter with the true model and 8192 particles tracks the belief on the same actions and observations.
Value networks with the same architecture as those of Tables~\ref{tab:results} and~\ref{tab:vc} are trained by squared error to predict the discounted return from each step, conditioned on either the mean and variance of the filter's particles (\textbf{Moments}) or the model's embedding (\textbf{VC / Pooled}).
We report the held-out $R^2$ under five-fold cross-validation over episodes, where each network trains on episodes from three folds and early stops with the fourth.
Table~\ref{tab:value-rep} shows that, for both models and at every training-set size, value networks conditioned on the embedding predict the return significantly more accurately than those conditioned on the moments (paired $t$-test over held-out episodes).
The moment estimates use 16 times as many particles, so a shortage of particles does not explain the gap.
This supports the attribution of BetaZero's gap on PocMan to conditioning on particle summary statistics, and conditioning on the combination of the pooled embedding and coupling summaries leads to better value estimates than conditioning on the pooled embedding alone.

%% file: sections/limitations.tex
\section{Limitations} \label{sec:limitations}
Paired simulations at the root spend the budget evenly among all actions, so each action's share shrinks as the action space grows.
Sequential halving discards half of the remaining root actions after each phase and assigns each unvisited action an estimate of the root's value, so $\pibar$ remains defined over every root action \citep{danihelka2022policy}.
Pairing the actions within each phase would keep the variance reduction of common sampled states and extend BETS to large action spaces.

BETS must be trained on each domain before it can search effectively, while AdaOPS, DESPOT, and POMCPOW have no training phase.
Against BetaZero, training improves returns where a particle set and its summary statistics represent the belief poorly, as on RockSample(20,20) and PocMan, but not on LightDark, where it adds computational overhead with no benefit over simpler belief models. 
Each simulation applies the NBF update, the policy network, and the value network once, so search takes roughly $K$ times as long per decision as the policy network alone (Appendix~\ref{app:wallclock}).
The policy network alone earns more return than every baseline does with search on RockSample(20,20) and both PocMan mazes (Table~\ref{tab:results}), so BETS can act without search where decisions must be fast.
Training also requires a simulator that exposes the realized hidden state, which might not always be available in practice.

%% file: sections/conclusion.tex
\section{Conclusion} \label{sec:conclusion}
This paper introduced Belief Embedding Tree Search (BETS), a Monte Carlo planner that represents the belief at search tree nodes with a generative distribution embedding.
The embedding lets the search sample states, and it gives the policy and value networks a learned compact representation of the belief to condition on.
We show that these function approximators benefit from conditioning on embeddings, rather than summaries of particle sets.
BETS is competitive or leading on every configuration, and even without search it outperforms every baseline by a large margin on RockSample(20,20) and PocMan, the configurations with the highest-dimensional beliefs.
Our experiments also suggest that adding per-variable coupling summaries to a GDE improves the model's representation on PocMan.

BETS shows that a learned belief embedding can take the place of a particle set inside a planner, and the same substitution is open to any algorithm that learns or plans on beliefs.
Embedding and updating beliefs is not specific to tree search, so the same ideas extend to other learning and planning algorithms for partially observable environments.
In games, player policies would enter the NBF update through the same particle propagation as the dynamics, so an embedding could represent public beliefs needed for many algorithms \citep{moravvcik2017deepstack,brown2020combining}.


%% file: appendix/domains.tex
\section{Domains} \label{app:domains}

\paragraph{RockSample.}
RockSample \citep{smith2004heuristic} places a rover on a square grid containing $k$ rocks at known positions, each independently good or bad.
The rover moves deterministically in the four cardinal directions, samples the rock at its current cell, or checks any rock with a noisy long-range sensor.
A check returns the rock's quality corrupted by noise whose accuracy decays exponentially with the distance to the rock.
Sampling a good rock earns a positive reward and turns the rock bad, sampling a bad rock is penalized, and the episode ends with a positive reward when the rover exits the east edge of the grid.
The rover's position is fully observable and transitions are deterministic.
We use the two configurations of \citet{moss2023betazero}: \textbf{RockSample(15,15)} is a $15 \times 15$ grid with 15 rocks and 20 actions, and \textbf{RockSample(20,20)} is a $20 \times 20$ grid with 20 rocks and 25 actions.
Both use a fixed rock layout, sensor efficiency 20, discount 0.95, and horizon 100.

\paragraph{LightDark.}
LightDark \citep{platt2010belief} is a continuous one-dimensional localization domain.
The agent's position is a real-valued state that is never observed directly.
Each step yields an observation of the position corrupted by Gaussian noise whose standard deviation grows with the distance from a light.
The actions move the agent one unit left or right or stop, and stopping ends the episode with a positive reward if the agent's true position is within distance 1 of the origin and a negative reward otherwise.
Both configurations draw the agent's start position from $\mathcal{N}(2, 3^2)$, and use a discount factor 0.9 and horizon 100.
\textbf{LightDark(5)} places the light at position 5.
The standard deviation of the noise is $|\state - 5| / \sqrt{2} + 0.01$ at position $\state$, and stopping earns $+10$ or $-10$.
\textbf{LightDark(10)} places the light at position 10 with noise standard deviation $|\state - 10| + 10^{-4}$, and stopping earns $+100$ or $-100$.

\paragraph{PocMan.}
PocMan \citep{silver2010monte} is a partially observable adaptation of the arcade game Pac-Man.
The agent navigates a maze, collecting food pellets while avoiding ghosts that chase it when nearby, and eating a power pill briefly lets the agent eat ghosts for a bonus.
The observation is 11 bits: sight of a ghost along each of the four cardinal directions, contact with walls in each direction, hearing a ghost within Manhattan distance 2, and smelling a pellet and smelling a power pill in the eight surrounding cells.
Rewards are $-1$ per step, $+10$ for a pellet or a power pill, $+25$ for eating a ghost, $-100$ for being caught, and $+1000$ for clearing the maze; the discount is 0.95.
\textbf{PocMan(M)} is a $9 \times 9$ maze with 59 reachable cells, three ghosts, three power pills, and horizon 100.
\textbf{PocMan(L)} is the $17 \times 19$ maze of \citet{silver2010monte} with 189 reachable cells, four ghosts, four power pills, and horizon 250.
In both mazes, the agent, the ghosts, and the power pills start on the same cells in every episode.
The hidden state is the ghosts' positions and headings together with the pellet field, in which each reachable cell holds a pellet independently with probability 0.5 at the start of an episode.
The simulator forbids a ghost from reversing its direction, so its heading constrains its next move.
The pellet field is not modelled by the generator and an observed pellet mask is given to all filters.

\paragraph{Policy and Value Network Inputs.}
The policy and value networks condition on the embedding $\bembed$ together with a vector $\obsfeat$ of features of the observed part of the state.
In every domain, $\obsfeat$ contains the fraction of the horizon elapsed.
On PocMan it also contains the fraction of pellets remaining, the power-pill timer, and the observed pellet mask over the reachable cells.

%% file: appendix/hyperparameters.tex
\section{Hyperparameters} \label{app:hyperparameters}

\paragraph{BETS.}
Search settings are shared between training and evaluation, and an annealed setting moves linearly from its first to its second value across the anneal window and keeps the second value afterwards.
An action node keeps at most 16 observation children, and once it is full, a traversal whose observation matches none of them descends to an existing child sampled in proportion to its visit count plus one.
On LightDark every sampled observation is new, so observation progressive widening lowers this limit to $\lceil k\, N(\bembed, \action)^{\alpha} \rceil$ (Table~\ref{tab:hyperparameters}).

\begin{table}[h]
\caption{BETS settings per configuration.
A range is annealed over the anneal window.}
\label{tab:hyperparameters}
\centering
\footnotesize
\setlength{\tabcolsep}{3.2pt}
\begin{tabular}{@{}lccccc@{}}
\toprule
 & \multicolumn{2}{c}{RockSample} & \multicolumn{2}{c}{LightDark} & PocMan \\
\cmidrule(lr){2-3} \cmidrule(lr){4-5}
 & (15,15) & (20,20) & (5) & (10) & (M), (L) \\
\midrule
\multicolumn{6}{@{}l}{\emph{Search}} \\
Simulations per decision $K$ & 600 & 600 & 99 & 99 & 100 \\
Particles per node $n$ & 256 & 256 & 256 & 256 & 512 \\
Exploration constant $c$ & \multicolumn{5}{c}{20 to 3} \\
Prior mix $\priormix$ & \multicolumn{5}{c}{0 to 0.75} \\
Temperature $\readtemp$ & 2 to 0.5 & 2 to 0.5 & 2 to 0.5 & 20 to 5 & 10 to 2.5 \\
Observation widening $(k, \alpha)$ & -- & -- & (4, 0.1) & (4, 0.1) & -- \\
\midrule
\multicolumn{6}{@{}l}{\emph{Training}} \\
Gradient steps & 150k & 300k & 75k & 150k & 60k \\
Anneal window (gradient steps) & 10k to 40k & 15k to 60k & 7.5k to 30k & 15k to 60k & 4k to 16k \\
Collectors $\times$ fragment length & 84 $\times$ 100 & 84 $\times$ 100 & 84 $\times$ 100 & 84 $\times$ 100 & 84 $\times$ 25 \\
Collection interval (gradient steps) & 400 & 400 & 400 & 400 & 100 \\
Initial collection rounds & 8 & 8 & 5 & 5 & 5 \\
Value target $\lambda$ & 1 & 1 & 1 & 1 & 0.95 \\
Optimizer & \multicolumn{5}{c}{Adam, learning rate $10^{-3}$, batch 32, gradient clip 1.0} \\
\midrule
\multicolumn{6}{@{}l}{\emph{Networks}} \\
Pooled embedding $\mu$ dimension & 32 & 32 & 32 & 32 & 64 \\
VC stream (width $\times$ layers) & 16 $\times$ 3 & 16 $\times$ 3 & -- & -- & 32 $\times$ 4 \\
Set transformer width & 128 & 128 & 128 & 128 & 256 \\
Set transformer blocks, inducing points & \multicolumn{5}{c}{3, 16} \\
Attention heads & \multicolumn{5}{c}{4} \\
Generator (layers $\times$ width) & AR, 3 $\times$ 48 & AR, 3 $\times$ 48 & FM, 3 $\times$ 64 & FM, 3 $\times$ 64 & AR, 3 $\times$ 48 \\
Value bins on $[f_{\min}, f_{\max}]$ & 66, $\pm 8.1$ & 66, $\pm 8.1$ & 66, $\pm 3$ & 66, $\pm 9.2$ & 128, $[-14, 38]$ \\
Policy and value networks & \multicolumn{5}{c}{MLP, 3 hidden layers of 128} \\
\bottomrule
\end{tabular}
\end{table}

\paragraph{BetaZero.}
BetaZero runs with the same number of simulations per decision and particles as BETS in every domain, and otherwise with its default settings on RockSample and LightDark.
On PocMan, BetaZero tracks the belief with a bootstrap particle filter.
Its networks condition on the mean and standard deviation of the player and ghost positions and the mean pellet and power-pill grids over its particles.
During policy iteration, BetaZero samples the root action from the visit counts weighted by the root action values ($z_q = z_n = 1$) at temperature $\tau$, where $\tau = 0$ takes the argmax, and at evaluation it takes the argmax under the weights $(z_q, z_n)$.
A dash for widening means it is disabled, so every action is expanded at once and each node keeps only the first belief it is reached with (Table~\ref{tab:bz-hyperparameters}).

\begin{table}[h]
\caption{BetaZero settings per configuration.
A cell with two values gives the value during policy iteration, then the value at evaluation.}
\label{tab:bz-hyperparameters}
\centering
\footnotesize
\setlength{\tabcolsep}{3pt}
\begin{tabular}{@{}lccccc@{}}
\toprule
 & \multicolumn{2}{c}{RockSample} & \multicolumn{2}{c}{LightDark} & PocMan \\
\cmidrule(lr){2-3} \cmidrule(lr){4-5}
 & (15,15) & (20,20) & (5) & (10) & (M), (L) \\
\midrule
\multicolumn{6}{@{}l}{\emph{Policy iteration}} \\
Iterations $\times$ episodes & 50 $\times$ 500 & 50 $\times$ 500 & 30 $\times$ 500 & 30 $\times$ 500 & 40 $\times$ 250 \\
Epochs per iteration & 10 & 10 & 50 & 50 & 10 \\
Optimizer & RMSProp & RMSProp & Adam & Adam & RMSProp \\
Learning rate & $10^{-3}$ & $10^{-3}$ & $10^{-4}$ & $10^{-4}$ & $10^{-4}$ \\
Value loss & MSE & MSE & MSE & MSE & MAE \\
Dropout & 0.5 & 0.5 & 0.2 & 0.2 & 0.5 \\
Hidden layers $\times$ width & 4 $\times$ 128 & 4 $\times$ 128 & 4 $\times$ 64 & 4 $\times$ 64 & 4 $\times$ 256 \\
Batch, $L_2$ penalty, samples & \multicolumn{5}{c}{1024, $10^{-5}$, $10^{5}$ from the latest iteration} \\
\midrule
\multicolumn{6}{@{}l}{\emph{Search}} \\
Simulations per decision & 100 / 600 & 100 / 600 & 100 / 99 & 100 / 99 & 100 \\
Particles & 1000 / 256 & 1000 / 256 & 500 / 256 & 500 / 256 & 512 \\
Exploration constant $c$ & 50 & 50 & 1 & 1 & 50 \\
Maximum depth & 15 & 4 & 10 & 10 & 10 \\
Action widening $(k_a, \alpha_a)$ & -- / (5, 0.9) & -- & (2, 0.25) & (2, 0.25) & -- \\
Belief widening $(k_b, \alpha_b)$ & -- / (1, 0) & -- / (1, 0) & (2, 0.1) & (2, 0.1) & -- \\
Temperature $\tau$ & 1 / 0 & 1.5 / 0 & 0 & 0 & 1 / 0 \\
Evaluation weights $(z_q, z_n)$ & (0.4, 0.9) & (0.5, 0.8) & (1, 1) & (1, 1) & (0, 1) \\
Bootstrapped $Q_0$ & no / yes & no / yes & no & no & no \\
\bottomrule
\end{tabular}
\end{table}

\paragraph{Online planners.}
AdaOPS, DESPOT, and POMCPOW run with the same number of simulations per decision and particles as BETS in every domain, and otherwise with the settings and heuristics from \citet{moss2023betazero} on RockSample and LightDark.
They use the preferred-actions rollout policy of \citet{silver2010monte} on PocMan.
The exit policy moves the rover east to the exit without sampling any rock, and on LightDark $r$ is the reward for a correct stop (Table~\ref{tab:planner-hyperparameters}).

\begin{table}[h]
\caption{AdaOPS, DESPOT, and POMCPOW settings per domain.}
\label{tab:planner-hyperparameters}
\centering
\footnotesize
\setlength{\tabcolsep}{4pt}
\begin{threeparttable}
\begin{tabular}{@{}lccc@{}}
\toprule
 & RockSample & LightDark & PocMan \\
\midrule
\multicolumn{4}{@{}l}{\emph{AdaOPS} \citep{wu2021adaptive}} \\
Lower, upper bound & exit policy, QMDP$^\dagger$ & $-r$, $r$ & -- \\
$m_{\min}$, $\delta$ & 100, 0.1 & 10, 1.0 & -- \\
Maximum depth & 90 & 90 & -- \\
\midrule
\multicolumn{4}{@{}l}{\emph{DESPOT} \citep{somani2013despot}} \\
Lower, upper bound & exit policy, QMDP$^\dagger$ & random policy, $r$ & preferred actions, fixed \\
Scenarios $K$, regularization $\lambda$ & 100, 0 & 30, 0.1 & 30, 0.01 \\
Maximum depth & \multicolumn{3}{c}{90} \\
\midrule
\multicolumn{4}{@{}l}{\emph{POMCPOW} \citep{sunberg2018online}} \\
Leaf value & exit policy$^\ddagger$ & MDP value & preferred-actions rollout \\
UCB constant & 10 & 10 & $\sqrt{2}$ \\
Action widening $(k_a, \alpha_a)$ & -- & (10, 0.5) & -- \\
Observation widening $(k_o, \alpha_o)$ & (1, 1) & (4, 0.03) & (1, 1) \\
Maximum depth & 90 & 20 & 100 \\
\bottomrule
\end{tabular}
\begin{tablenotes}
\item[$^\dagger$] Fixed upper bound $V_{\max} = 138.3$ on RockSample(20,20).
\item[$^\ddagger$] Random rollouts on RockSample(20,20).
\end{tablenotes}
\end{threeparttable}
\end{table}

%% file: appendix/ar-conditioning.tex
\section{Generator Architectures}\label{app:generators}

\subsection{Autoregressive Transformer}\label{app:ar-conditioning}
The generator of Section~\ref{sec:vc} is a causal transformer that keeps one token per variable of the sample, and token $j$ emits the logits of the factor $p_\flowparams(x_j \mid x_{<j}, \bembed)$.
Each token's input is the sum of three terms: a learned positional embedding for token $j$, an embedding of the preceding value $x_{j-1}$ with a learned start token in place of $x_0$ at $j = 1$, and a conditioning vector computed from $\bembed$.

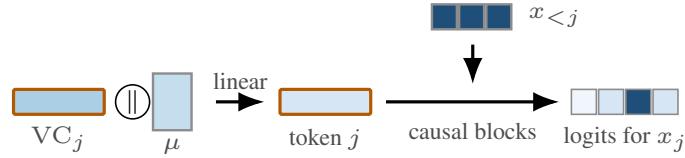
\begin{figure}[t]
  \centering
  \scalebox{1.5}{\input{fig/ar-conditioning.tikz}}
  \caption{\textbf{Per-variable conditioning of the autoregressive generator.}
  Token $j$ combines the coupling summary $\mathrm{VC}_j$ of variable $j$ with the pooled embedding $\mu$ shared by every token.
  The causal blocks attend to the preceding values $x_{<j}$, and the output is the logits of the factor $p_\flowparams(x_j \mid x_{<j}, \bembed)$.}
  \label{fig:ar-conditioning}
\end{figure}

Pooled conditions the generator on one vector.
A single linear map sends $\bembed$ to the token dimension and the result is added to every token identically.
The conditioning is the same at every token.
VC instead splits $\bembed$ into the pooled embedding $\mu$ and the $D$ per-variable coupling summaries, and applies one linear map, shared by every token, to the concatenation $[\mu \,;\, \mathrm{VC}_j]$ at token $j$ (Figure~\ref{fig:ar-conditioning}).
Token $j$ and summary $j$ index the same variable, so factor $j$ sees the shared pooled statistics and the summary for the variable it predicts.

\subsection{Flow-Matching} \label{app:fm}
LightDark's hidden state is a real-valued position, so its generator is a conditional flow-matching model \citep{lipman2022flow,tong2023conditional}.
A velocity field, an MLP with three hidden layers of width 64 that takes the current sample, the flow time, and $\bembed$ as input, transports noise to a sample of the belief in 50 integration steps.
Its belief loss is the conditional flow-matching regression at the realized state, and its embedding uses the pooled stream.

%% file: fig/ar-conditioning.tikz
\begin{tikzpicture}[
    >=Latex,
    every node/.style={font=\small},
    /utils/exec=\definecolor{mplBluesA}{HTML}{F0F6FC}%
        \definecolor{mplBluesB}{HTML}{D6E6F4}%
        \definecolor{mplBluesC}{HTML}{ABD0E6}%
        \definecolor{mplBelief}{HTML}{1F4E79},
    tokenbar/.style={
        draw, semithick, rounded corners=0.5pt, fill=mplBluesC,
        minimum width=8mm, minimum height=2.2mm, inner sep=0pt,
    },
    tokenbarl/.style={tokenbar, fill=mplBluesB},
    hl/.style={draw=orange!70!black, thick},
    summarys/.style={
        draw=black!45, semithick, sharp corners, fill=mplBluesC!70,
        minimum width=3.4mm, minimum height=9mm, inner sep=1pt,
    },
    summaryzoom/.style={summarys, minimum height=5mm},
    lbl/.style={font=\scriptsize, inner sep=1pt, align=center},
    sub/.style={font=\tiny, inner sep=1pt, align=center, text=black!70},
    arr/.style={->, thick, shorten >=1pt, shorten <=1pt},
]


\node[tokenbar, hl] (ci) at (5mm, 0mm) {};
\node[sub] at (5mm, -3.3mm) {$\mathrm{VC}_j$};
\node[draw, thin, circle, fill=white, inner sep=0pt, minimum size=3mm, font=\tiny] at (11.5mm, 0mm) {$\|$};

\node[summaryzoom] (muz) at (15mm, 0mm) {};
\node[sub] at (15mm, -3.9mm) {$\mu$};

\draw[arr] (18.5mm, 0mm) -- (23.5mm, 0mm);
\node[sub] at (21mm, 1.8mm) {linear};
\node[tokenbarl, hl] (toki) at (28.5mm, 0mm) {};
\node[sub] at (28.5mm, -3.3mm) {token $j$};

\begin{scope}[shift={(41.5mm,7.5mm)}]
    \foreach \c in {0,1,2} {
        \draw[black!45, semithick, fill=mplBelief]
            ($(-3.6mm,-1.1mm) + (\c*2.4mm,0)$) rectangle ++(2.2mm,2.2mm);
    }
\end{scope}
\node[sub, anchor=west] at (46mm, 7.5mm) {$x_{<j}$};
\draw[arr] (41.5mm, 5.4mm) -- (41.5mm, 1.2mm);

\draw[arr] (33.5mm, 0mm) -- (49.5mm, 0mm);
\node[sub] at (41.5mm, -2.6mm) {causal blocks};
\begin{scope}[shift={(55mm,0mm)}]
    \foreach \c in {0,1,2,3} {
        \ifnum\c=2 \def\cc{mplBelief}\else
            \pgfmathtruncatemacro{\s}{mod(\c,2)}
            \ifcase\s \def\cc{mplBluesA}\else \def\cc{mplBluesB}\fi
        \fi
        \draw[black!45, thin, fill=\cc]
            ($(-4.8mm,-1.1mm) + (\c*2.4mm,0)$) rectangle ++(2.2mm,2.2mm);
    }
\end{scope}
\node[sub] at (55mm, -3.3mm) {logits for $x_j$};

\end{tikzpicture}

%% file: appendix/wallclock.tex
\section{Wall-Clock Time} \label{app:wallclock}
We measure the wall clock time used by BETS, averaged over 100 episodes.
Standard errors are below 3\% of the mean.
Each simulation applies the NBF update, the policy network, and the value network once, so search takes roughly $K$ times as long per decision as the policy network alone (Table~\ref{tab:wallclock}).
The policy network alone, which earns more return than every baseline on RockSample(20,20) and both PocMan mazes (Table~\ref{tab:results}), takes under 25 ms per decision on every configuration.

\begin{table}[h]
\caption{Milliseconds per decision of BETS on one NVIDIA GeForce RTX 5080.}
\label{tab:wallclock}
\centering
\footnotesize
\begin{threeparttable}
\begin{tabular}{@{}lcrrrr@{}}
\toprule
 & & \multicolumn{3}{c}{search} & \\
\cmidrule(lr){3-5}
Domain & $n$ & $K = 50$ & $K = 100$ & $K = 200$ & policy \\
\midrule
RockSample(15,15) & 256 & 140 & 229 & 460 & 2.8 \\
 & 512 & 248 & 410 & 812 & 4.5 \\
RockSample(20,20) & 256 & 176 & 349 & 688 & 4.0 \\
 & 512 & 323 & 641 & 1289 & 6.6 \\
LightDark(5) & 256 & 237 & 454 & 901 & 5.6 \\
 & 512 & 452 & 866 & 1722 & 9.7 \\
LightDark(10) & 256 & 238 & 455 & 902 & 5.6 \\
 & 512 & 452 & 867 & 1724 & 9.7 \\
PocMan(M) & 256 & 253 & 517 & 1042 & 5.6 \\
 & 512 & 425 & 879 & 1729 & 9.2 \\
PocMan(L) & 256 & 894 & 1876 & 3722 & 19.6 \\
 & 512 & 1107 & 2320 & 4692 & 24.1 \\
\bottomrule
\end{tabular}
\begin{tablenotes}
\item Each $K$ is rounded up to a multiple of $|\actionspace|$, so that every round of the search selects every root action once.
\end{tablenotes}
\end{threeparttable}
\end{table}